\documentclass[lettersize,journal]{IEEEtran}
\usepackage{amsmath,amsfonts}
\usepackage{algorithmic}
\usepackage{algorithm}
\usepackage{array}
\usepackage[caption=false,font=normalsize,labelfont=sf,textfont=sf]{subfig}
\usepackage{textcomp}
\usepackage{stfloats}
\usepackage{url}
\usepackage{verbatim}
\usepackage{graphicx}
\usepackage{cite}
\usepackage{mathrsfs}
\usepackage{threeparttable}
\usepackage{multirow}
\usepackage{booktabs}
\usepackage[pagebackref=true,breaklinks=true,colorlinks,bookmarks=false,citecolor=blue,linkcolor=blue,urlcolor=blue]{hyperref}
\newcommand{\email}[1]{\href{mailto:}{#1}} 

\begin{document}

\title{FedKLPR: KL-Guided Pruning-Aware Federated Learning for Person Re-Identification}

\author{Po-Hsien Yu, Yu-Syuan Tseng, and Shao-Yi Chien, \textit{Member, IEEE}
\thanks{Manuscript created April 5, 2026;}
\thanks{P.-H. Yu, Y.-S Tseng and S.-Y. Chien are with the Media IC and System
Lab, the Graduate Institute of Electronics Engineering and Department
of Electrical Engineering, National Taiwan University, Taipei 106319, Taiwan.\\E-mail: \email{michaelyu@media.ee.ntu.edu.tw}, \email{ystseng@media.ee.ntu.edu.tw}, and \email{sychien@ntu.edu.tw}}
}

\markboth{IEEE Transactions on Multimedia}%
{Shell \MakeLowercase{\textit{et al.}}: A Sample Article Using IEEEtran.cls for IEEE Journals}


\maketitle

\begin{abstract}
Person re-identification (re-ID) is a fundamental task in intelligent surveillance and public safety. Federated learning (FL) provides a privacy-preserving paradigm for collaborative model training without centralized data collection. However, deploying FL in real-world re-ID systems remains challenging due to statistical heterogeneity caused by non-IID client data and the substantial communication overhead incurred by frequent transmission of large-scale models. To address these challenges, we propose FedKLPR, a lightweight and communication-efficient federated learning framework for person re-ID. FedKLPR consists of three key components. First, KL-Divergence-Guided training, including the KL-Divergence Regularization Loss (KLL) and KL-Divergence-aggregation Weight (KLAW), is introduced to mitigate statistical heterogeneity and improve convergence stability under non-IID settings. Second, unstructured pruning is incorporated to reduce communication overhead, and the Pruning-ratio-aggregation Weight (PRAW) is proposed to measure the relative importance of client parameters after pruning. Together with KLAW, PRAW forms KL-Divergence-Prune Weighted Aggregation (KLPWA), enabling effective aggregation of pruned local models under heterogeneous data distributions. Third, Cross-Round Recovery (CRR) adaptively controls pruning across communication rounds to prevent excessive compression and preserve model accuracy. Experiments on eight benchmark datasets demonstrate that FedKLPR achieves substantial communication savings while maintaining competitive accuracy. Compared with state-of-the-art methods, FedKLPR reduces communication cost by 40\%--42\% on ResNet-50 while achieving better overall performance.
\end{abstract}


\section{Introduction}
\IEEEPARstart{P}{erson} re-identification (re-ID) aims to identify individuals across non-overlapping camera networks under variations in viewpoint, illumination, and occlusion. It has been widely studied for surveillance and smart city applications, where pedestrian matching across distributed cameras is required. Recent advances in unsupervised person re-ID have reduced the reliance on labor-intensive identity annotations \cite{pre_id_cvpr, render_re_id, mutual_mean-teaching, reid_spatial_cvpr, dis_iccv, iics, dy_iccv, un_reid_iccv, un_reid_eccv, un_reid_arxiv, 27, CAP}. However, most existing methods still rely on centralized training, which raises privacy and scalability concerns in real-world deployments.

Federated learning (FL) \cite{FedSH, FedSea, FedMRSG, PDKD, FedAvg} provides a decentralized training paradigm that preserves data privacy by keeping raw data on local clients. In FL, each client trains a local model using its private dataset, and the central server updates the global model by aggregating local parameters. The updated global model is then redistributed to clients for subsequent local optimization. Since raw data are not transmitted to the server or shared among clients, FL has been increasingly applied to privacy-sensitive tasks, including person re-ID. For example, FedReID \cite{FedReIDBench} introduced federated training for person re-ID and achieved promising performance while preserving data privacy. However, FedReID relies on labeled data, whose collection is costly and labor-intensive in practical surveillance scenarios. To reduce the dependency on annotations, several unsupervised federated re-ID methods have been proposed \cite{FedUReID, FedUCC, FedUCC2, FedUCA, FedCAPR}. For instance, FedUReID \cite{FedUReID} jointly optimized cloud and edge models to mitigate statistical heterogeneity.

\begin{figure}[!t]
\begin{center}
   \includegraphics[width=0.9\linewidth]{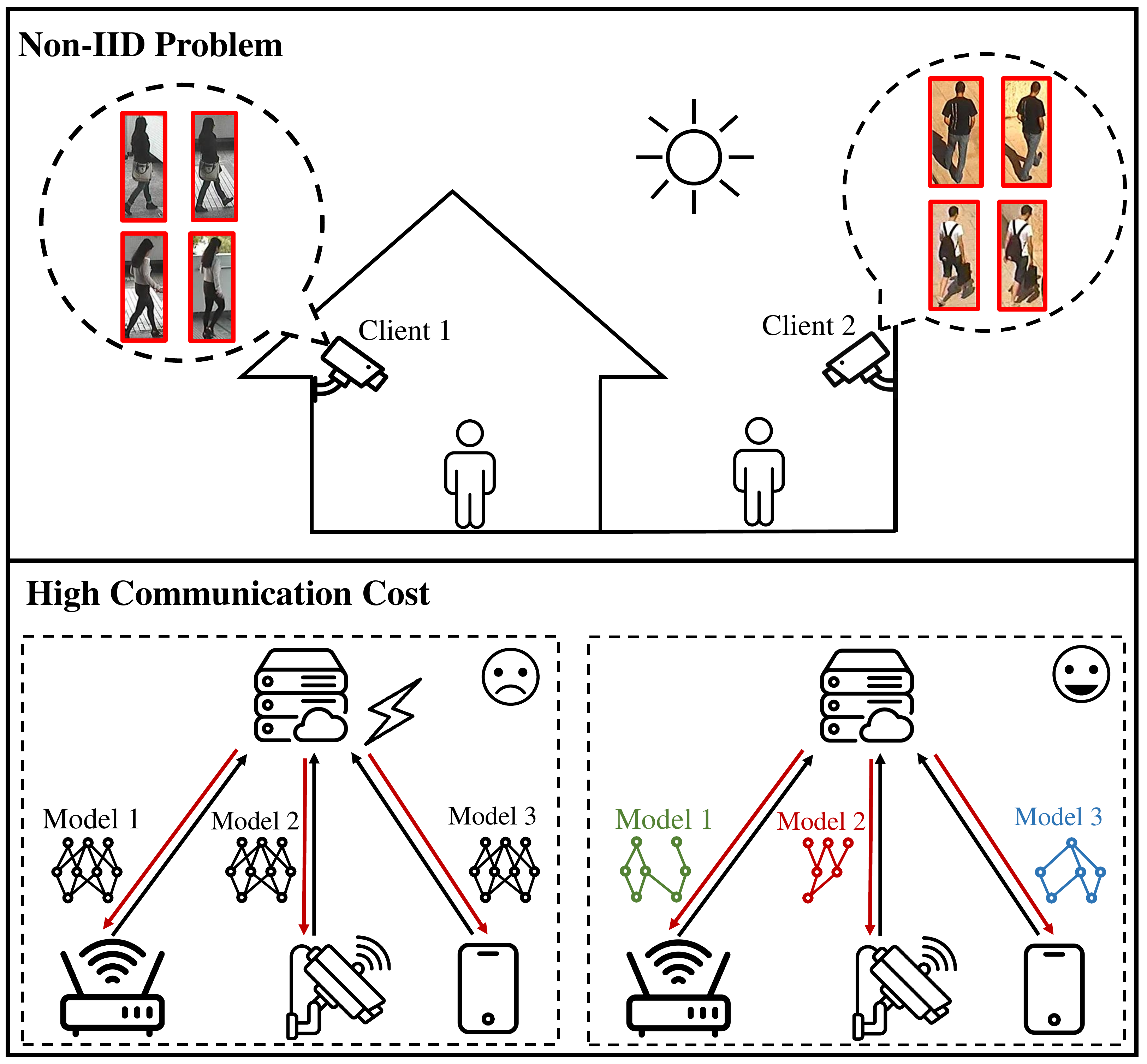}
   \vspace{-10pt}
   \caption{Two major challenges in unsupervised federated learning re-ID systems: non-IID data variances across clients and the communication cost between clients and the cloud server.}
\label{fig:challenge}
\end{center}
\end{figure}

Nonetheless, existing unsupervised federated re-ID methods still face two major challenges, as illustrated in Fig.~\ref{fig:challenge}. First, non-IID data distributions across clients, including label skew, feature skew, and quantity skew, can lead to feature misalignment and degrade model performance \cite{FedSea}. Although FedCAPR \cite{FedCAPR} introduced Identity-Distributed Equalization and Cosine Similarity Regularization to reduce inter-client distribution discrepancies, its performance remains limited on small-scale datasets. Second, federated re-ID systems incur substantial communication costs because model parameters must be repeatedly exchanged between clients and the central server. Model pruning is a promising approach to reduce communication overhead, and several methods \cite{SubFedAvg, LotteryFL, FedDIP} have incorporated pruning into FL. However, aggressive pruning may remove parameters that encode discriminative visual cues, while insufficient pruning provides limited communication savings. Existing pruning-based FL methods often lack adaptive mechanisms to balance compression and recognition accuracy under heterogeneous client distributions.

To address these challenges, we propose FedKLPR, a federated person re-ID framework that integrates KL-divergence-guided optimization with adaptive model pruning. Built upon FedCAPR \cite{FedCAPR}, FedKLPR replaces cosine-similarity regularization with Kullback--Leibler divergence (KL divergence) \cite{kl_divergence} to model distributional discrepancies between local and global representations. During local training, the proposed KL-Divergence Regularization Loss (KLL) aligns the output probability distributions of the locally updated model and the initial personalized model to improve convergence stability. During aggregation, the KL-Divergence-aggregation Weight (KLAW) adaptively weights client updates according to the KL divergence between the initial personalized model and the locally trained model, thereby emphasizing informative client updates under non-IID data distributions.
To further reduce communication overhead, FedKLPR incorporates unstructured pruning into local training. We introduce the Pruning-ratio-aggregation Weight (PRAW) to reflect the relative importance of retained parameters after pruning. By jointly integrating KLAW and PRAW, we develop KL-Divergence-Prune Weighted Aggregation (KLPWA), a pruning-aware aggregation strategy that accounts for both distributional discrepancy and client-specific sparsity. In addition, we propose the Cross-Round Recovery (CRR) mechanism to prevent excessive performance degradation caused by aggressive pruning. Experiments on eight benchmark datasets, including DukeMTMC \cite{DukeMTMC}, Market1501 \cite{Market_1501}, iLIDS-VID \cite{iLIDS}, CUHK03 \cite{CUHK03}, Prid2011 \cite{PRID}, VIPeR \cite{VIPeR}, CUHK01 \cite{CUHK01}, and 3DPeS \cite{3DPeS}, demonstrate the effectiveness of FedKLPR in improving recognition performance while reducing communication costs in federated re-ID.

In summary, the major contributions of the FedKLPR are as follows:

\begin{itemize}
    \item KL-Divergence-Guided training: We introduce KL-Divergence-Guided training, which consists of KL-Divergence Regularization Loss (KLL) and KL-Divergence-aggregation Weight (KLAW), to improve local training and global aggregation under heterogeneous client distributions. This design enables FedKLPR to better capture informative non-IID updates and enhance aggregation robustness.
    
    \item KL-Divergence-Prune Weighted Aggregation (KLPWA): 
    We propose KL-Divergence-Prune Weighted Aggregation (KLPWA), which integrates the Pruning-ratio-aggregation Weight (PRAW) with KLAW to form a unified pruning-aware aggregation strategy. By jointly considering distributional discrepancy and client-specific sparsity, KLPWA improves aggregation effectiveness for pruned local models in non-IID federated settings.

    \item  Cross-Round Recovery (CRR):
    We propose a Cross-Round Recovery (CRR) mechanism to regulate iterative pruning across communication rounds. CRR adaptively determines whether pruning should be applied, thereby reducing unnecessary accuracy degradation while maintaining model compactness.
    
\end{itemize}

The remainder of this paper is organized as follows. Section II reviews related work on unsupervised federated person re-ID and federated pruning methods. Section III presents the proposed FedKLPR framework. Section IV reports the experimental results and analyses. Section V concludes this paper.

\section{Related Work}
\subsection{Unsupervised Federated Person re-ID}
Unsupervised federated learning has recently emerged as a promising paradigm for person re-identification (re-ID), as it addresses both data privacy concerns and the need for diverse training distributions. In unsupervised re-ID, existing methods can be broadly categorized into domain-adaptive and fully unsupervised approaches. Domain-adaptive methods \cite{pre_id_cvpr, render_re_id, mutual_mean-teaching} leverage labeled source-domain data to improve generalization to unlabeled target domains. For example, Lv \textit{et al.} \cite{reid_spatial_cvpr} showed that transferring spatial-temporal features can help mitigate domain gaps. In contrast, fully unsupervised methods \cite{dis_iccv, iics} rely entirely on unlabeled data and commonly adopt clustering-based, graph-based \cite{dy_iccv, un_reid_iccv}, or k-NN-based strategies \cite{un_reid_eccv, un_reid_arxiv} to generate pseudo-labels. Among clustering-based methods, DBSCAN \cite{density_aaai} and BUC \cite{26} are widely used due to their ability to progressively refine pseudo-labels and improve training reliability \cite{27}. In addition, the CAP framework \cite{CAP} introduces a camera-aware proxy loss and a contrastive loss to reduce intra-camera and inter-camera variations.

Although unsupervised person re-ID methods have achieved promising performance, their conventional centralized training paradigm raises privacy concerns. Federated learning (FL) provides a decentralized alternative by allowing clients to retain data locally and share only model updates with a central server. Recent studies have explored the integration of federated learning with unsupervised person re-ID to address privacy constraints and cross-client data diversity. FedUCC \cite{FedUCC} proposes a coarse-to-fine framework that includes broad information extraction, BatchNorm-based personalized representation \cite{batchnorm}, patch-level feature refinement, and feature decomposition to alleviate cross-camera divergence. However, its multi-stage training process requires up to 100 communication rounds, leading to considerable communication overhead. FedUCA \cite{FedUCA} incorporates the CAP framework \cite{CAP} into federated learning using EasyFL \cite{easyfl}, highlighting the importance of camera-level variation modeling in unsupervised federated re-ID. FedCAPR, proposed by Tseng \textit{et al.} \cite{FedCAPR}, further improves unsupervised federated re-ID by addressing clustering, aggregation, and data heterogeneity. Nevertheless, existing unsupervised federated re-ID methods largely overlook model size and communication efficiency. Since these methods do not incorporate model compression, they require repeated transmission of full model parameters during communication rounds. This limitation becomes a major deployment bottleneck for bandwidth-limited or resource-constrained edge devices, where both communication efficiency and model scalability are essential.

\begin{figure*}[!t]
\begin{center}
   \includegraphics[width=0.9\linewidth]{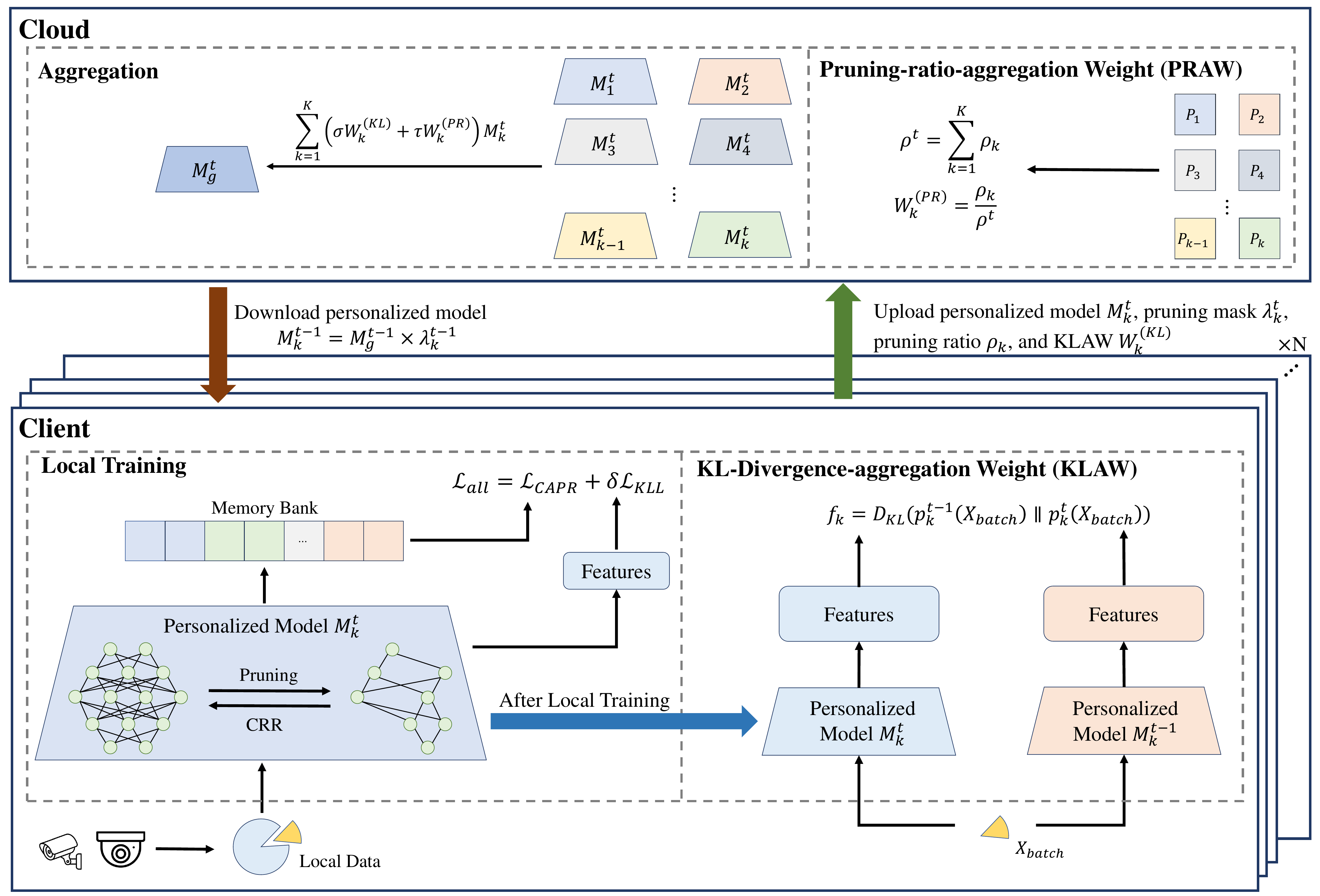}
   \vspace{-10pt}
   \caption{Overview of the FedKLPR framework, consisting of a cloud server and eight clients; each client performs local training with KLL, applies unstructured pruning with CRR, computes KLAW, and uploads the local model, pruning mask, pruning ratio, and KLAW to the cloud, where PRAW is calculated. The cloud combines KLAW and PRAW to aggregate the local models.}
\label{fig:fedklpr}
\end{center}
\end{figure*}
\subsection{Pruning in Federated Learning}
The transmission of large and redundant model parameters poses a major bottleneck in distributed learning systems \cite{fl_challenge}. To address this issue, model compression and sparsification techniques have been widely investigated to reduce communication overhead among distributed nodes. Compared with gradient sparsification \cite{gradient}, reducing the model size is particularly important in federated learning (FL), as it can simultaneously decrease communication cost, storage demand, and inference latency, thereby improving the practicality of FL on resource-constrained devices. Existing model compression techniques, including weight pruning, quantization, low-rank factorization, transferred convolutional filters, and knowledge distillation \cite{kn_distill}, have been extensively studied in centralized learning. In this work, we focus on weight pruning for communication-efficient FL.

Several pruning-based methods have recently been proposed to improve the efficiency of FL by reducing computational and communication costs. LotteryFL \cite{LotteryFL} performs iterative pruning on client devices, but it lacks an adaptive pruning strategy, which may limit its efficiency under heterogeneous settings. PruneFL \cite{prunefl} applies unstructured pruning to a coarsely pruned model to further reduce local computation. However, it adopts a uniform sparsity pattern across clients, limiting its flexibility in adapting to heterogeneous data distributions and device capabilities. SubFedAvg \cite{SubFedAvg} performs personalized pruning for each client, allowing local models to better adapt to client-specific data. Nevertheless, its pruning decision mainly relies on immediate accuracy feedback and does not sufficiently consider whether the model has recovered its pre-pruning performance before further pruning, which may lead to performance degradation.
Recent studies have also explored server-side pruning. FedDIP \cite{FedDIP} addresses non-IID data by introducing a regularization penalty during training. However, it distributes a single pruned model to all clients, which limits its ability to adapt to client-specific data characteristics. In contrast, our method performs adaptive pruning on the client side, enabling each client to dynamically adjust model complexity according to its local data distribution and computational capacity. This design improves personalization while reducing communication and computation costs in heterogeneous FL environments.

\section{Methodology}
In this section, we present the overall FedKLPR framework, which consists of three stages: KL-Divergence-Guided training, KL-Divergence-Prune Weighted Aggregation (KLPWA), and Cross-Round Recovery (CRR). First, KL-Divergence Regularization Loss (KLL) and KL-Divergence-aggregation Weight (KLAW) are introduced to improve training stability and aggregation robustness under heterogeneous client distributions. Second, the Pruning-ratio-aggregation Weight (PRAW) is computed to measure the relative importance of pruned client models and is jointly integrated with KLAW to update the global model through KLPWA. Finally, CRR adaptively regulates pruning decisions across communication rounds to mitigate accuracy degradation caused by aggressive pruning.

\subsection{Overall Architecture}

Fig.~\ref{fig:fedklpr} illustrates the overall architecture of FedKLPR, which consists of a cloud server and multiple edge clients. FedKLPR is built upon FedCAPR~\cite{FedCAPR} and inherits its unsupervised local re-ID training pipeline, including the memory bank mechanism and the camera-aware contrastive learning loss $\mathscr{L}_{\text{capr}}$. On top of this baseline, FedKLPR introduces KL-Divergence-Guided training, KL-Divergence-Prune Weighted Aggregation (KLPWA), and Cross-Round Recovery (CRR) to improve aggregation robustness under non-IID client distributions and reduce communication overhead through adaptive pruning.

During KL-Divergence-Guided training, each client downloads its personalized model from the cloud server. The personalized model is derived from the global model using the client-specific pruning mask $\lambda^{t-1}_k$, which selects parameters according to local characteristics. Each client then trains the personalized model on its private dataset. To mitigate excessive model drift caused by heterogeneous data distributions, we introduce the KL-Divergence Regularization Loss (KLL), which constrains the discrepancy between the locally trained model and the initial personalized model. Meanwhile, each client updates its local memory bank with feature embeddings to maintain temporal consistency. A subset of local samples is further fed into both the initial personalized model and the locally trained model to obtain their output distributions. The KL divergence between these distributions is used to compute the KL-Divergence-aggregation Weight (KLAW), which measures the distributional change induced by local training. The computed KLAW is transmitted to the server and used for global aggregation.

During aggregation, each client also sends its pruning ratio to the cloud server. The server incorporates these pruning ratios into the aggregation-weight calculation to obtain the Pruning-ratio-aggregation Weight (PRAW), which accounts for the relative importance of retained parameters in each local model. KLAW and PRAW are then jointly integrated to form the proposed KL-Divergence-Prune Weighted Aggregation (KLPWA) scheme for global model updating. By considering both distributional discrepancy and client-specific sparsity, KLPWA enables the server to aggregate local models while preserving the structural characteristics induced by local pruning. The updated global model is then redistributed to all clients and personalized using their corresponding pruning masks.

To further reduce communication overhead, FedKLPR applies unstructured pruning within the local training phase. The proposed Cross-Round Recovery (CRR) strategy determines whether pruning should be activated in the current round based on performance feedback from previous rounds. This mechanism provides adaptive pruning control and helps prevent excessive accuracy degradation. Through the iterative process of local training, aggregation-weight computation, weighted aggregation, and cross-round pruning supervision, FedKLPR balances local adaptability and global consistency under both statistical and structural heterogeneity.

\subsection{KL-Divergence-Guided Training} 
In federated learning, KL-Divergence-Guided training begins by distributing the current global model from the cloud server to each client. Each client updates the received model using its private data and returns the updated parameters to the server for aggregation, enabling collaborative learning while preserving data privacy. In FedKLPR, we introduce KL-Divergence Regularization Loss (KLL) and KL-Divergence-aggregation Weight (KLAW) to improve training stability under heterogeneous data distributions. KLL constrains excessive local model drift, while KLAW assigns adaptive aggregation weights based on the distributional discrepancy between the initial personalized model and the locally trained model. The two mechanisms are detailed as follows.
\textbf{\textit{1) KL-Divergence Regularization Loss (KLL).}}
In federated learning, local training is independently performed on each client using private data, which may lead to model divergence under statistical heterogeneity. Large discrepancies among client models can destabilize global aggregation and weaken knowledge integration. To address this issue, we introduce the Kullback--Leibler divergence regularization loss (KLL) during local training. KLL constrains the divergence between the locally updated model and the initial personalized model received from the server, improving representation consistency and training stability. The output distributions used in KLL are defined as
\begin{equation}
\label{eq:softmax_logits}
\begin{split}
p_k^t(x) = \text{softmax}(M^t_k(x)), 
\end{split}
\end{equation}
and the KLL term is formulated as
\begin{equation}
\label{eq:kll}
\mathscr{L}_{\text{KLL}} = \frac{1}{|X|}\sum_{x\in X}D_{KL}(p^{t-1}_k(x) \parallel p^t_k(x)), 
\end{equation}
where $x$ denotes an image sample from the local dataset $X$, $M_k^t$ is the locally updated model of the $k$-th client at communication round $t$, and $M_k^{t-1}$ denotes the initial personalized model broadcast by the server at the beginning of round $t$. Moreover, $p_k^t$ denotes the output probability distribution produced by the locally updated model, while $p_k^{t-1}$ denotes that produced by the initial personalized model.

The KL divergence between the probability distributions obtained from their logits via softmax quantifies the distributional shift induced by local training. 
By minimizing the KL divergence during training, FedKLPR encourages alignment between the output probability distributions of the locally updated model and the initial personalized model, thereby mitigating excessive model drift and improving optimization stability.
This regularization is particularly useful in federated learning, where heterogeneous data distributions can cause substantial variations in local model updates. The overall loss function of FedKLPR is defined as
\begin{equation}
\label{eq:overall_loss}
\mathscr{L}_{\text{all}} = \mathscr{L}_{\text{capr}} + \delta \mathscr{L}_{\text{KLL}},
\end{equation}
where $\mathscr{L}_{\text{capr}}$ denotes the camera-aware loss in FedCAPR, and $\delta$ is the weighting coefficient of the KLL term.

\textbf{\textit{2) KL-Divergence-aggregation Weight (KLAW).}}
In addition to local regularization, KL divergence is further used to guide global aggregation. Under non-IID data distributions, different clients may exhibit different degrees of update-induced distributional change after local training. To account for this variation, we introduce the KL-Divergence-aggregation Weight (KLAW), which assigns aggregation weights according to the discrepancy between the initial personalized model and the locally trained model.

As illustrated in Fig.~\ref{fig:fedklpr}, after receiving the personalized model $M^{t-1}{k}$ from the server, the $k$-th client feeds a reference batch $X_{batch}$ into $M^{t-1}{k}$ to obtain the initial output distribution $p^{t-1}_k(X_{batch})$. After local training, the same batch is fed into the updated local model $M^t_k$ to obtain $p^t_k(X_{batch})$. The client then computes the KL-based discrepancy score as

\begin{equation}
\label{eq:klw_client}
f_k = D_{KL}(p^{t-1}_k(X_{batch})\parallel p^t_k(X_{batch})).
\end{equation}
A larger $f_k$ indicates a greater distributional change between the initial personalized model and the locally updated model. The KL-based aggregation weight is then obtained by normalizing $f_k$ across all clients:
\begin{equation}
    \label{eq:klaw}
    W_k^{(KL)} = \frac{f_k}{f}, \text{ where } f = \sum^K_{k=1} f_k.
\end{equation}
The global model is updated by aggregating the $K$ local models as
\begin{equation}
\label{eq:klw_global}
M^t_g = \sum^K_{k=1}W_k^{(KL)}M^t_k.
\end{equation}
where $f$ is the normalization factor. Through KLAW, FedKLPR adaptively adjusts the contribution of each client during aggregation according to its KL-based distributional change, thereby improving aggregation robustness under heterogeneous client distributions.

\subsection{Pruning-ratio-aggregation Weight and Aggregation} 
In federated learning, repeated transmission of model parameters between clients and the server incurs substantial communication overhead. To reduce this cost, pruning is applied to compress local models before aggregation. However, different clients may adopt different pruning ratios, leading to structural heterogeneity among local models. To address this issue, we propose the Pruning-ratio-aggregation Weight (PRAW), which accounts for client-specific pruning ratios during aggregation. We further introduce KL-Divergence-Prune Weighted Aggregation (KLPWA), which combines PRAW with KLAW to update the global model by jointly considering distributional discrepancy and pruning-induced sparsity. The details of PRAW and KLPWA are presented below.

\textbf{\textit{1) Pruning-ratio-aggregation Weight (PRAW).}}
A higher pruning ratio indicates that fewer parameters are retained on the client side, and the remaining parameters are expected to be more critical to the local task. If this pruning-induced importance is ignored during aggregation, retained parameters may be overly affected by updates from other clients, leading to suboptimal global model optimization. To address this issue, we introduce the Pruning-ratio-aggregation Weight (PRAW), which uses the client-specific pruning ratio to adjust the contribution of each local model during aggregation. The PRAW of the $k$-th client is defined as
\begin{equation}
\label{eq:prw}
W_k^{(PR)} = \frac{\rho_k}{\rho^t}, 
\quad
\rho^t = \sum^K_{k=1}\rho_k,
\end{equation}
where $\rho_k$ denotes the pruning ratio of the $k$-th client, $W_k^{(PR)}$ is its corresponding PRAW, and $\rho^t$ is the normalization factor over all $K$ clients. In this way, clients with higher pruning ratios are assigned larger aggregation weights, allowing the aggregation process to better preserve the influence of highly compressed local models.

\textbf{\textit{2) KL-Divergence-Prune Weighted Aggregation (KLPWA).}}
Once KLAW and PRAW are obtained, they are integrated to form the final aggregation weight of KLPWA. The global model is updated as
\begin{equation}
\label{eq:klpwa}
M_g^t = \sum^K_{k=1}
\left(\sigma W_k^{(KL)} + \tau W_k^{(PR)}\right) M^t_k,
\end{equation}
where $W_k^{(KL)}$ and $W_k^{(PR)}$ denote the KLAW and PRAW of the $k$-th client, respectively. The hyperparameters $\sigma$ and $\tau$ control the relative contributions of the KL-divergence-based weight and the pruning-ratio-based weight, with $\sigma + \tau = 1$. KLAW captures the distributional discrepancy between the initial personalized model and the locally trained model, while PRAW accounts for the pruning-induced structural importance of each local model. By jointly considering distributional change and client-specific sparsity, KLPWA enables more effective aggregation of pruned local models under heterogeneous federated settings.

\begin{figure}[!t]
\begin{center}
   \includegraphics[width=0.9\linewidth]{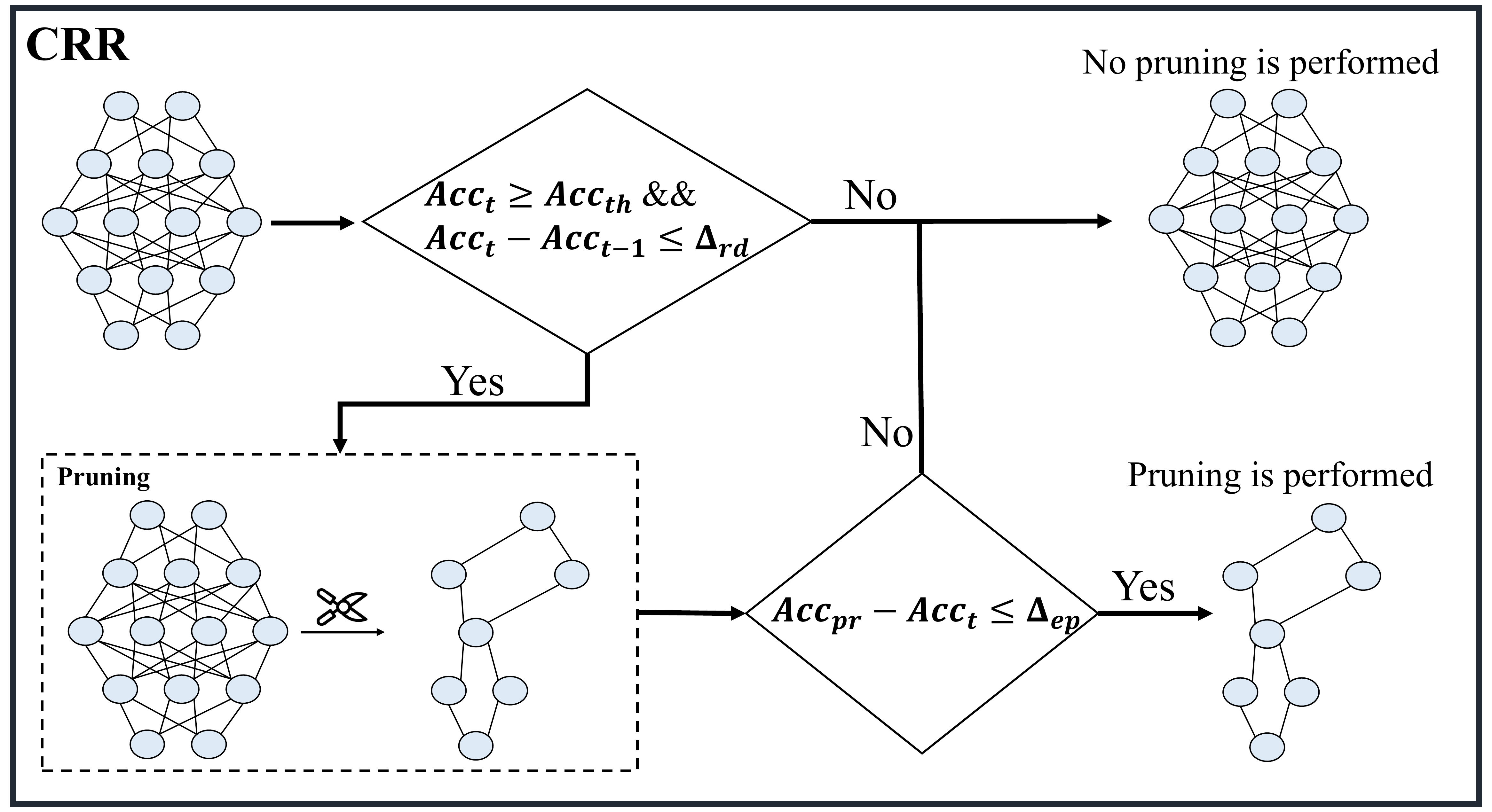}
   \vspace{-10pt}
   \caption{The CRR mechanism with two-stage verification: Stage 1 checks whether the training accuracy surpasses the target accuracy $Acc_{th}$ and the improvement over previous rounds is within $\Delta_{rd}$; Stage 2 ensures the accuracy drop before and after pruning is less than $\Delta_{ep}$; pruning is executed only if both conditions are satisfied.}
\label{fig:crr}
\end{center}
\end{figure}

\subsection{Cross-Round Recovery (CRR)}
In federated learning, local client models are periodically aggregated to update the global model, which is then redistributed to clients for subsequent local training. However, when the number of local epochs is limited, a client's accuracy may decrease after aggregation compared with the previous communication round. Applying pruning under such conditions can further reduce model capacity and amplify performance degradation, leading to progressively weaker local models. To address this issue, we propose Cross-Round Recovery (CRR), a two-stage pruning supervision mechanism that allows the model to recover through intermediate retraining before further pruning is applied. Unlike conventional pruning methods that evaluate performance degradation within a single round, CRR additionally considers cross-round performance trends caused by federated aggregation, thereby improving pruning reliability and stabilizing local model accuracy during iterative compression.

Fig.~\ref{fig:crr} illustrates the CRR procedure, which determines whether pruning should be performed in the current round. In Stage 1, pruning is enabled only when the training accuracy exceeds the target threshold $Acc_{th}$ and the inter-round accuracy degradation is smaller than $\Delta_{rd}$. In Stage 2, after tentative pruning, the accuracy drop between the pre-pruned and pruned models must be smaller than $\Delta_{ep}$. Pruning is retained only when both conditions are satisfied; otherwise, the pruning operation is skipped or revoked for the current local training round.
\section{Experimental Results}
In this section, we evaluate FedKLPR under two settings: non-pruning federated person re-identification and pruning-based federated learning. We first describe the experimental setup and compare FedKLPR with recent non-pruning federated re-ID methods, including FedUReID \cite{FedUReID}, FedUCC \cite{FedUCC}, FedUCC+ \cite{FedUCC2}, FedUCA \cite{FedUCA}, and FedCAPR \cite{FedCAPR}. We then compare FedKLPR with representative pruning-based FL methods, including LotteryFL \cite{LotteryFL}, SubFedAvg \cite{SubFedAvg}, and FedDIP \cite{FedDIP}. Finally, ablation studies are conducted to examine the effectiveness of each proposed component.

\begin{table}[ht]
\caption{
Statistics of the eight benchmark re-ID datasets.
}
\vspace{-5pt}
\label{tab:datasets}
\centering
\footnotesize
\tabcolsep=3pt
\begin{tabular}{l c c c c c}
\toprule[2pt]
\multirow{2}{*}{Dataset} & \multirow{2}{*}{Cam IDs} & \multicolumn{2}{c} {Training Set} & \multicolumn{2}{c}{Testing Set} \\
 & & IDs No. & Imgs No. & Query No. & Gallery No. \\
\midrule[0.3pt]
DukeMTMC\cite{DukeMTMC}  & 8 & 702 & 16522 & 2228 & 17611 \\
Market1501\cite{Market_1501} & 6 & 751 & 12936 & 3368 & 19732 \\
iLIDS-VID\cite{iLIDS}             & 2 & 59  & 248   & 98   & 130   \\
CUHK03\cite{CUHK03}           & 2 & 767 & 7365  & 1400 & 5332 \\
Prid2011\cite{PRID}               & 2 & 285 & 3744  & 100  & 649 \\
VIPeR\cite{VIPeR}             & 2 & 316 & 632   & 316  & 316 \\
CUHK01\cite{CUHK01}           & 2 & 485 & 1940  & 972  & 972 \\
3DPeS\cite{3DPeS}             & 2 & 93  & 450   & 246  & 316 \\
\bottomrule[2pt]
\end{tabular}
\end{table}
\subsection{Experimental Setup}
\textbf{\textit{1) Datasets.}} Following prior works \cite{FedUReID, FedUCC, FedUCC2, FedUCA, FedCAPR}, we adopt a federated learning setting with eight benchmark person re-identification (re-ID) datasets, including DukeMTMC \cite{DukeMTMC}, Market1501 \cite{Market_1501}, iLIDS-VID \cite{iLIDS}, CUHK03 \cite{CUHK03}, Prid2011 \cite{PRID}, VIPeR \cite{VIPeR}, CUHK01 \cite{CUHK01}, and 3DPeS \cite{3DPeS}, where each dataset is treated as an independent client without data sharing. These datasets contain diverse variations in camera number, dataset scale, identity distribution, and appearance, as summarized in Table~\ref{tab:datasets}.

\textbf{\textit{2) Network Architecture.}} We adopt FedCAPR \cite{FedCAPR} as the baseline framework and use ResNet-50 pretrained on ImageNet \cite{imagenet} as the backbone feature extractor. For the non-pruning evaluation, following state-of-the-art federated unsupervised person re-ID methods \cite{FedUReID, FedUCA, FedUCC, FedUCC2, FedCAPR}, we evaluate the effectiveness of KL-Divergence Regularization Loss (KLL) and KL-Divergence-aggregation Weight (KLAW) under non-IID data distributions. For the pruning evaluation, we use the same backbone to assess the proposed pruning-related components, including Pruning-ratio-aggregation Weight (PRAW) and Cross-Round Recovery (CRR), in terms of accuracy preservation and communication efficiency.

\textbf{\textit{3) Implementation Details.}} 
FedKLPR was instantiated with eight clients, each corresponding to one benchmark re-ID dataset. Training was conducted for 20 global communication rounds, with five local epochs per client in each round. We used the Adam optimizer with a base learning rate of $\eta = 3.5 \times 10^{-4}$. The camera-aware loss coefficient followed FedCAPR, and the KLL regularization coefficient $\delta$ was empirically set to 0.15. For KLPWA, $\sigma$ and $\tau$ were set to 0.8 and 0.2, respectively, in the pruning setting, and to 1.0 and 0.0 in the non-pruning setting.

After local training, clients with accuracy above 55\% entered the personalized adaptive pruning phase. The pruning controller removed up to 9\% of parameters per communication round. CRR used two thresholds, $\Delta_{rd}$ and $\Delta_{ep}$, both set to 1.5\%, to regulate pruning. Pruning was activated when the inter-round accuracy degradation was below $\Delta_{rd}$, and was retained only when the post-pruning accuracy drop remained within $\Delta_{ep}$; otherwise, the pruning operation was revoked. The pruning process was performed for up to 10 evaluation epochs and terminated once the target pruning ratio of 70\% or the maximum number of evaluation epochs was reached.

\textbf{\textit{4) Evaluation Metrics.}} 
Person re-identification (re-ID) is formulated as an image retrieval task, where the objective is to retrieve gallery images sharing the same identity as a given query image. Following standard evaluation protocols in prior works \cite{FedCAPR}, we adopt two widely used metrics: mean Average Precision (mAP) and Rank-1 accuracy from the Cumulative Matching Characteristic (CMC) curve. These metrics evaluate retrieval quality and top-ranked identification accuracy.

\begin{table*}[ht]
\caption{Rank-1 accuracy (\%) and mAP (\%) comparison of federated unsupervised re-ID methods on ResNet-50}
\vspace{-5pt}
\label{tab:non_pruning_sota_comparison}
\centering
\footnotesize
\tabcolsep=3pt
\begin{threeparttable}
\begin{tabular}{l c c c c c c c c c c c c c c c c}
\toprule[2pt]
\multirow{2}{*}{Methods} & \multicolumn{2}{c}{DukeMTMC} & \multicolumn{2}{c} {Market1501} & \multicolumn{2}{c}{iLIDS-VID} & \multicolumn{2}{c}{CUHK03} & \multicolumn{2}{c}{Prid2011}
    & \multicolumn{2}{c}{VIPeR} & \multicolumn{2}{c}{CUHK01} & \multicolumn{2}{c}{3DPeS} \\

& Rank-1 & mAP & Rank-1 & mAP & Rank-1 & mAP & Rank-1 & mAP & Rank-1 & mAP & Rank-1 & mAP & Rank-1 & mAP & Rank-1 & mAP  \\
\midrule[0.3pt]
FedUReID \cite{FedUReID} & 51.0\% & -    & 65.2\% & -    & 73.5\% & -    & 8.9\%  & -    & 38.0\% & -    & 26.6\% & -    & 43.6\% & -    & 65.5\% & -    \\
FedUCC   \cite{FedUCC}   & 78.8\% & 60.5\% & 86.5\% & 65.5\% & 74.7\% & 59.7\% & 9.6\%  & 9.7\%  & 58.9\% & 63.1\% & 31.3\% & 36.7\% & 78.3\% & 75.3\% & 68.9\% & 50.9\% \\
FedUCC+  \cite{FedUCC2}  & -    & -    & 90.3\% & 75.2\% & 82.8\% & 72.0\% & 38.7\% & 35.5\% & 69.0\% & 72.0\% & 43.0\% & 48.6\% & 80.1\% & 76.6\% & 73.2\% & 57.8\% \\
FedUCA   \cite{FedUCA}   & 81.0\% & 66.5\% & \textbf{92.5\%} & 79.4\% & 80.5\% & -    & 50.0\% & -    & 75.5\% & -    & 51.0\% & -    & 86.0\% & -    & 85.0\% & -    \\

FedCAPR  \cite{FedCAPR}  & 82.5\% & 69.8\% & 92.3\% & \textbf{81.6\%} & 79.6\% & 78.4\% & 71.4\% & 67.5\% & 83.0\% & 87.9\% & 67.1\% & 73.4\% & \textbf{95.6\%} & \textbf{95.1\%} & \textbf{86.2\%} & 80.9\% \\

\midrule[0.3pt]

FedKLPR$^*$ (Ours)              & \textbf{83.2\%} & \textbf{69.9\%} & \textbf{92.5\%} & 81.4\% & \textbf{84.7\%} & \textbf{79.3\%} & \textbf{71.6\%} & \textbf{68.9\%} & \textbf{85.0\%} & \textbf{88.3\%} & \textbf{68.4\%} & \textbf{75.5\%} & 94.1\% & 94.5\% & 85.4\% & \textbf{81.3\%} \\

\bottomrule[2pt]
\end{tabular}
    \begin{tablenotes}
        \item[*] This FedKLPR framework only adopts the KLL and KLAW methods.
    \end{tablenotes}
\end{threeparttable}
\end{table*}
\subsection{Performance Evaluation}
Table~\ref{tab:non_pruning_sota_comparison} presents the comparison between FedKLPR and state-of-the-art methods under the non-pruning setting. Compared with FedUReID \cite{FedUReID}, FedUCC \cite{FedUCC}, FedUCC+ \cite{FedUCC2}, and FedUCA \cite{FedUCA}, FedKLPR achieves clear performance improvements on most benchmark datasets. In particular, compared with FedUCC+, FedKLPR improves Rank-1 accuracy by 1.9\% on iLIDS-VID, 32.9\% on CUHK03, 16.0\% on Prid2011, 25.4\% on VIPeR, 14.0\% on CUHK01, and 12.2\% on 3DPeS. For mAP, FedKLPR improves performance by 7.3\% on iLIDS-VID, 33.4\% on CUHK03, 16.3\% on Prid2011, 26.9\% on VIPeR, 17.9\% on CUHK01, and 23.5\% on 3DPeS. Compared with FedUCA, FedKLPR further improves Rank-1 accuracy by 4.2\% on iLIDS-VID, 21.6\% on CUHK03, 9.5\% on Prid2011, 17.4\% on VIPeR, and 8.1\% on CUHK01.

Compared with FedCAPR \cite{FedCAPR}, which serves as the baseline framework, FedKLPR achieves competitive overall performance and shows clear advantages on several challenging datasets. Specifically, FedKLPR improves Rank-1 accuracy by 5.1\% on iLIDS-VID and 2.0\% on Prid2011. On VIPeR, FedKLPR outperforms FedCAPR by 1.3\% in Rank-1 accuracy and 2.1\% in mAP. On the remaining datasets, FedKLPR maintains performance comparable to FedCAPR, indicating that the proposed KL-divergence-guided design improves robustness under heterogeneous client distributions without degrading performance on relatively stable benchmarks. 

\begin{table}[ht]
\caption{Accuracy (\%) and communication cost comparison of FedKLPR and existing federated pruning methods on small-scale datasets. \textbf{Boldface} indicates the highest value, while \underline{underlining} denotes the second-highest value.}
\vspace{-5pt}
\label{tab:resnet50_prune_sota_comparison_small}
\centering
\footnotesize
\tabcolsep=3pt
\begin{threeparttable}
\begin{tabular}{l l c c c c}
\toprule[2pt]
Datasets & Methods & Rank-1 & mAP &  PR$^*$ & CC$^{**}$ \\
\midrule[0.3pt]
\midrule[0.3pt]

\multirow{4}{*}{iLIDS-VID} & LotteryFL \cite{LotteryFL}  & 71.43\%          & 65.31\%          & 53.20\% & 2.33GB \\
& SubFedAvg \cite{SubFedAvg}  & \underline{78.57\%} & 75.28\% & 70.01\% & 2.27GB \\
& FedDIP \cite{FedDIP} & \textbf{80.61\%} & \underline{78.45\%} & 70.03\% & 1.98GB \\
& FedKLPR (Ours) & \textbf{80.61\%} & \textbf{78.58\%} & 70.01\% & 2.09GB \\
\midrule[0.3pt]

\multirow{4}{*}{Prid2011} & LotteryFL & 72.00\%          & 79.58\%          & 45.20\% & 2.26GB \\
& SubFedAvg  & \underline{79.00\%} & \underline{84.48\%} & 62.46\% & 2.35GB \\
& FedDIP & 77.00\% & 83.22\% & 66.34\% & 2.37GB \\
& FedKLPR (Ours)  & \textbf{84.00\%} & \textbf{86.94\%} & 70.01\% & 2.13GB \\
\midrule[0.3pt]

\multirow{4}{*}{VIPeR} & LotteryFL  & 37.97\%          & 48.15\%          & 51.14\% & 2.27GB \\
& SubFedAvg  & \textbf{65.82\%} & \textbf{72.68\%} & 68.05\% & 2.27GB \\
& FedDIP   & 60.76\% & 68.50\% & 70.01\% & 2.01GB \\
& FedKLPR (Ours)   & \underline{64.56\%} & \underline{72.26\%} & 67.85\% & 2.18GB \\
\midrule[0.3pt]

\multirow{4}{*}{3DPeS} & LotteryFL  & 74.80\%          & 62.55\%          & 42.83\% & 2.25GB \\
& SubFedAvg & 84.55\% & 75.95\% & 70.01\% & 2.23GB \\
& FedDIP  & \textbf{85.77\%} & \underline{77.86\%} & 70.03\% & 1.99GB \\
& FedKLPR (Ours) & \underline{84.96\%} & \textbf{78.39\%} & 70.01\% & 2.09GB \\

\bottomrule[2pt]
\end{tabular}
    \begin{tablenotes}
        \item[*] PR represents the pruning ratio of the final model on the client side.
        \item[**] Communication Cost (CC) is measured over the course of 20 global communication rounds. 
    \end{tablenotes}
\end{threeparttable}
\end{table}
\begin{table}[ht]
\caption{Comparison of accuracy (\%) and communication cost among different federated learning methods, including FedAvg \cite{FedAvg}, FedProx \cite{fedprox}, and FedCAPR \cite{FedCAPR}, using ResNet-50 on small-scale datasets. \textbf{Boldface} indicates the highest value, while \underline{underlining} denotes the second-highest value.}
\vspace{-5pt}
\label{tab:resnet50_baseline_comparison}
\centering
\footnotesize
\tabcolsep=3pt
\begin{threeparttable}
\begin{tabular}{l l c c c c}
\toprule[2pt]
Datasets & Methods & Rank-1 & mAP & PR$^*$ & CC$^{**}$ \\
\midrule[0.3pt]
\midrule[0.3pt]

\multirow{4}{*}{iLIDS-VID} & FedAvg & 77.55\%  & 75.04\%          & 70.01\% & 2.15GB \\

& FedProx ($\mu$ = 0.1) & \textbf{80.61\%} & \underline{77.05\%} & 53.17\% & 2.57GB \\
& FedCAPR & \underline{78.57\%}  & 76.87\%          & 70.01\% & 2.18GB \\
& FedKLPR (Ours) & \textbf{80.61\%}  & \textbf{78.58\%}          & 70.01\% & 2.09GB \\

\midrule[0.3pt]
\multirow{4}{*}{Prid2011} & FedAvg  & 77.00\%  & 82.48\%          & 36.70\% & 2.72GB \\
& FedProx ($\mu$ = 0.1) & 80.00\% & 85.07\% & 55.24\% & 2.44GB \\
& FedCAPR & \underline{83.00\%}           & \textbf{87.00\%}          & 45.31\% & 2.50GB \\
& FedKLPR (Ours) & \textbf{84.00\%}  & \underline{86.94\%} & 70.00\% & 2.13GB \\

\midrule[0.3pt]
\multirow{4}{*}{VIPeR} & FedAvg  & \textbf{65.82\%}  & \textbf{72.68\%}          & 52.69\% & 2.46GB \\

& FedProx ($\mu$ = 0.1) & 62.03\% & 68.90\% & 52.20\% & 2.53GB \\
& FedCAPR & 63.29\%           & 71.36\%          & 64.32\% & 2.25GB \\

& FedKLPR (Ours) & \underline{64.56\%}  & \underline{72.26\%} & 67.85\% & 2.18GB \\

\midrule[0.3pt]
\multirow{4}{*}{3DPeS} & FedAvg  & 82.11\%  & \underline{77.73\%}          & 70.01\% & 2.07GB \\

& FedProx ($\mu$ = 0.1) & \underline{82.93\%} & 76.39\% & 63.23\% & 2.45GB \\
& FedCAPR & 81.71\%           & 76.47\%          & 70.01\% & 2.08GB \\
& FedKLPR (Ours) & \textbf{84.96\%}  & \textbf{78.39\%} & 70.01\% & 2.09GB \\

\bottomrule[2pt]
\end{tabular}
    \begin{tablenotes}
        \item[*] PR represents the pruning ratio of the final model on the client side. 
        \item[**] Communication Cost (CC) is measured over the course of 20 global communication rounds. 
    \end{tablenotes}
\end{threeparttable}
\end{table}

\subsection{Pruning Evaluation}
To validate the effectiveness of the pruning strategies proposed in FedKLPR, we conduct comparative experiments against state-of-the-art federated pruning methods on small-scale datasets. For each dataset, the target pruning ratio is set to 70\%, and we evaluate the pruned models in terms of Rank-1 accuracy, mAP, pruning ratio, and communication cost. The results of these comparisons are summarized in Table~\ref{tab:resnet50_prune_sota_comparison_small}. In Table~\ref{tab:resnet50_prune_sota_comparison_small}, FedKLPR significantly reduces the communication cost on small-scale datasets, lowering the transmission volume to only 58.0\%--60.3\% of the original amount. Meanwhile, in terms of performance, FedKLPR achieves competitive or superior performance compared with existing pruning-based FL methods \cite{LotteryFL, SubFedAvg, FedDIP} on these datasets, while maintaining a favorable balance between pruning ratio and communication cost. Compared with LotteryFL \cite{LotteryFL}, FedKLPR consistently delivers superior accuracy across all small-scale datasets. In particular, FedKLPR improves the Rank-1 accuracy by 9.18\% and the mAP by 13.27\% on iLIDS-VID, by 12.00\% and 7.36\% on Prid2011, and by 10.16\% and 15.84\% on 3DPeS. Moreover, LotteryFL exhibits limited pruning effectiveness on VIPeR, achieving a pruning ratio of only 51.14\%. By contrast, FedKLPR reaches a much higher pruning ratio of 67.85\% on the same dataset, while still improving the Rank-1 accuracy by 26.59\% and the mAP by 24.11\%.



Furthermore, FedKLPR achieves better performance than SubFedAvg \cite{SubFedAvg} on most datasets. Specifically, FedKLPR improves Rank-1 accuracy by 2.04\% and mAP by 3.30\% on iLIDS-VID, and further improves mAP by 2.44\% on 3DPeS. On Prid2011, FedKLPR outperforms SubFedAvg by 5.00\% in Rank-1 accuracy and 2.46\% in mAP. In addition, FedKLPR reaches the target pruning ratio of 70\%, whereas SubFedAvg achieves only 62.46\%, yielding a 7.54\% higher pruning ratio. These results indicate that FedKLPR provides a better balance between communication efficiency and recognition performance across different datasets. Compared with FedDIP \cite{FedDIP}, which achieves a client-side pruning ratio of up to 70\%, FedKLPR shows better accuracy preservation on challenging datasets. In particular, FedKLPR improves Rank-1 accuracy by 7.00\% and mAP by 3.72\% on Prid2011. On VIPeR, FedKLPR improves Rank-1 accuracy by 3.80\% and mAP by 3.76\%.

\begin{table*}
\caption{Rank-1 accuracy ablation study on ResNet-50. The \underline{underline} is the highest accuracy obtained without applying unstructured pruning technique, whereas the \textbf{bold} denotes the highest measurements with unstructured pruning.}
\vspace{-5pt}
\label{tab:resnet50_record_init_ablation_study}
\centering
\footnotesize
\tabcolsep=3pt
\begin{threeparttable}
\begin{tabular}{c c c c c | c c c c c c c c }
\toprule[2pt]
\multirow{2}{*}{KLL} & \multirow{2}{*}{KLAW} & \multirow{2}{*}{P$^*$} & \multirow{2}{*}{CRR} & \multirow{2}{*}{PRAW} & DukeMTMC & Market1501 & iLIDS-VID & CUHK03 & Prid2011 & VIPeR & CUHK01 & 3DPeS \\
&&&&& Rank-1 / PR$^{**}$ & Rank-1 / PR & Rank-1 / PR & Rank-1 / PR & Rank-1 / PR & Rank-1 / PR & Rank-1 / PR & Rank-1 / PR \\
\midrule[0.3pt]
&&&& & 82.5 / 0.0  & 92.3 / 0.0 & 79.6 / 0.0 & 71.4 / 0.0 & 83.0 / 0.0 & 67.1 / 0.0 & \underline{95.6} / 0.0 & \underline{86.2} / 0.0 \\

\checkmark &&&&& 82.9 / 0.0 & \underline{92.8} / 0.0 & 79.6 / 0.0 & \underline{71.6} / 0.0 & 83.0 / 0.0 & 65.8 / 0.0 & 95.1 / 0.0 & 84.6 / 0.0 \\

\checkmark & \checkmark & &&& \underline{83.2} / 0.0 & 92.5 / 0.0 & \underline{84.7} / 0.0 & \underline{71.6} / 0.0 & \underline{85.0} / 0.0 & \underline{68.4} / 0.0 & 94.1 / 0.0 & 85.4 / 0.0 \\
\midrule[0.3pt]
\midrule[0.3pt]

\checkmark & \checkmark & \checkmark &&& \textbf{82.9} / 70.0 & \textbf{93.0} / 70.0 & \textbf{80.6} / 70.0 & 72.5 / 67.4 & 78.0 / 70.0 & 62.3 / 69.3 & 94.4 / 70.0 & 80.9 / 70.0 \\

\checkmark & \checkmark & \checkmark & \checkmark & & 82.6 / 70.0 & 92.8 / 70.0 & 77.6 / 27.6 & \textbf{73.2} / 67.4 & 81.0 / 27.2 & \textbf{67.4} / 65.9 & \textbf{94.9} / 70.0 & 82.1 / 70.0 \\

\checkmark & \checkmark & \checkmark & \checkmark & \checkmark & 
82.5 / 70.0 & 92.4 / 70.0 & \textbf{80.6} / 70.0 & 72.1 / 66.8 & \textbf{84.0} / 70.0 & 64.6 / 67.9 & 94.4 / 70.0 & \textbf{85.0} / 70.0 \\

\bottomrule[2pt]
\end{tabular}
    \begin{tablenotes}
        \item[*] P denotes the unstructured pruning method.
        \item[**] PR represents the pruning ratio of the final model on the client side.
    \end{tablenotes}
\end{threeparttable}
\end{table*}
\subsection{Comparison with Various Federated Learning Methods}
We further evaluate representative federated learning methods, including FedAvg \cite{FedAvg}, FedProx \cite{fedprox}, and FedCAPR \cite{FedCAPR}, under the pruning setting. For a fair comparison, the target pruning ratio is fixed at 70\% for all methods. We examine whether each method can reach the predefined pruning target and evaluate the final model performance. Table~\ref{tab:resnet50_baseline_comparison} summarizes the comparison between FedKLPR and the baseline FL methods. In Table~\ref{tab:resnet50_baseline_comparison}, FedKLPR consistently ranks among the top two methods across all small-scale datasets. Compared with FedAvg \cite{FedAvg}, FedKLPR improves Rank-1 accuracy by 3.06\% and mAP by 3.54\% on iLIDS-VID. On Prid2011, FedKLPR improves Rank-1 accuracy by 7.00\% and mAP by 4.46\%. On 3DPeS, FedKLPR improves Rank-1 accuracy by 2.85\% and mAP by 0.66\%. Moreover, on VIPeR, FedKLPR achieves performance comparable to FedAvg while attaining a pruning ratio of 67.85\%, which is 15.16\% higher than that of FedAvg. Similarly, on Prid2011, FedKLPR reaches a pruning ratio of 70.00\%, whereas FedAvg achieves only 36.70\%, yielding a 33.30\% higher pruning ratio. These results indicate that FedKLPR provides a better balance between recognition performance and pruning capability than FedAvg under the same target pruning setting.

Moreover, Table~\ref{tab:resnet50_baseline_comparison} shows that FedKLPR achieves performance comparable to FedProx on iLIDS-VID while attaining a 16.84\% higher pruning ratio. On the remaining small-scale datasets, including Prid2011, VIPeR, and 3DPeS, FedKLPR consistently outperforms FedProx in both recognition performance and pruning capability. On Prid2011, FedKLPR improves Rank-1 accuracy by 4.00\%, mAP by 1.87\%, and pruning ratio by 14.76\%. On VIPeR, FedKLPR improves Rank-1 accuracy by 2.53\%, mAP by 3.36\%, and pruning ratio by 15.65\%. On 3DPeS, FedKLPR further improves Rank-1 accuracy by 2.03\%, mAP by 2.00\%, and pruning ratio by 6.78\%. These results indicate that FedKLPR provides a better balance between recognition performance and pruning effectiveness than FedProx.

Furthermore, FedKLPR achieves comparable or better performance than FedCAPR while obtaining substantially higher pruning ratios on several datasets. Although both methods reach a pruning ratio of 70\% on iLIDS-VID and 3DPeS, FedKLPR improves Rank-1 accuracy by 2.04\% and mAP by 1.71\% on iLIDS-VID. On 3DPeS, FedKLPR further improves Rank-1 accuracy by 3.25\% and mAP by 1.92\%. On Prid2011 and VIPeR, FedKLPR achieves pruning ratios that are 24.69\% and 3.53\% higher than those of FedCAPR. Notably, although FedKLPR prunes more parameters on these two datasets, it maintains performance comparable to FedCAPR.

\subsection{Ablation Studies}
We conduct an ablation study with the ResNet-50 backbone to evaluate the contribution of each component in FedKLPR. The results are reported in Table~\ref{tab:resnet50_record_init_ablation_study}. By introducing KL-Divergence Regularization Loss (KLL), the discrepancy between local and global models is constrained, which stabilizes global aggregation. KLL improves performance by approximately 0.5\% on large-scale datasets, including DukeMTMC, Market1501, and CUHK03. However, since KLL makes local updates more conservative, slight performance degradation is observed on smaller-scale datasets, with Rank-1 accuracy drops of 1.3\% on VIPeR and 1.6\% on 3DPeS. When KLAW is further incorporated to reweight client updates according to KL-based distributional change, FedKLPR achieves more notable improvements on challenging datasets, including 5.1\% on iLIDS-VID, 2.0\% on Prid2011, 2.6\% on VIPeR, and 0.8\% on 3DPeS. Meanwhile, no clear degradation is observed on stronger clients, indicating that the proposed KL-divergence-guided design improves weaker clients while preserving the performance of stronger ones.

When unstructured pruning is applied to reduce communication cost, noticeable accuracy degradation is observed, particularly on smaller-scale datasets. The Rank-1 accuracy decreases by 4.1\% on iLIDS-VID, 7.0\% on Prid2011, 6.1\% on VIPeR, and 4.5\% on 3DPeS. To mitigate this degradation, we introduce Cross-Round Recovery (CRR) and Pruning-ratio-aggregation Weight (PRAW). As shown in Table~\ref{tab:resnet50_record_init_ablation_study}, CRR prevents continued pruning when cross-round accuracy degradation is detected, thereby reducing the progressive performance loss caused by excessive pruning. CRR improves Rank-1 accuracy by 3.0\% on Prid2011, 5.1\% on VIPeR, and 1.2\% on 3DPeS, while also providing modest gains on CUHK03 and CUHK01. Since CRR restricts pruning to preserve accuracy, it also lowers the achievable pruning ratio, with reductions of 42.4\% on iLIDS-VID, 42.8\% on Prid2011, and 3.4\% on VIPeR.

To compensate for this limitation, PRAW is further incorporated into the aggregation stage. The results show that PRAW maintains comparable Rank-1 accuracy on most datasets while improving both accuracy and pruning capability on several challenging benchmarks. After integrating PRAW, the Rank-1 accuracy on iLIDS-VID and Prid2011 is further improved by 3.0\%, and the pruning ratio on both datasets reaches 70\%. On 3DPeS, PRAW also improves Rank-1 accuracy by 2.9\%. Overall, these results demonstrate that combining CRR and PRAW provides a more favorable balance between model accuracy and pruning ratio across heterogeneous clients.

\section{Conclusion}
In this paper, we proposed FedKLPR, a federated learning framework for privacy-preserving person re-identification (re-ID), to address two key challenges in federated re-ID systems: convergence instability caused by statistical heterogeneity and high communication overhead in resource-constrained edge environments. FedKLPR integrates three main components. First, KL-Divergence-Guided training introduces KL-Divergence Regularization Loss (KLL) and KL-Divergence-aggregation Weight (KLAW). KLL constrains the discrepancy between the output probability distributions of the locally updated model and the initial personalized model to improve convergence stability, while KLAW adaptively weights local models according to their KL-based distributional change during aggregation. Second, Pruning-ratio-aggregation Weight (PRAW) and KL-Divergence-Prune Weighted Aggregation (KLPWA) are proposed to support pruning-aware aggregation. PRAW accounts for the relative importance of retained parameters after pruning, and KLPWA integrates KLAW and PRAW to improve global aggregation under non-IID data distributions and client-specific sparsity. Third, Cross-Round Recovery (CRR) mitigates performance degradation caused by iterative pruning through a two-stage cross-round verification mechanism. Experimental results on eight benchmark datasets demonstrate the effectiveness of FedKLPR in both non-pruning and pruning settings. In particular, FedKLPR achieves strong performance on small-scale datasets and reduces communication cost by approximately 40\%--42\% with ResNet-50 while maintaining competitive or better accuracy than existing state-of-the-art methods.

\begin{IEEEbiography}[{\includegraphics[width=1in,height=1.25in, clip,keepaspectratio]{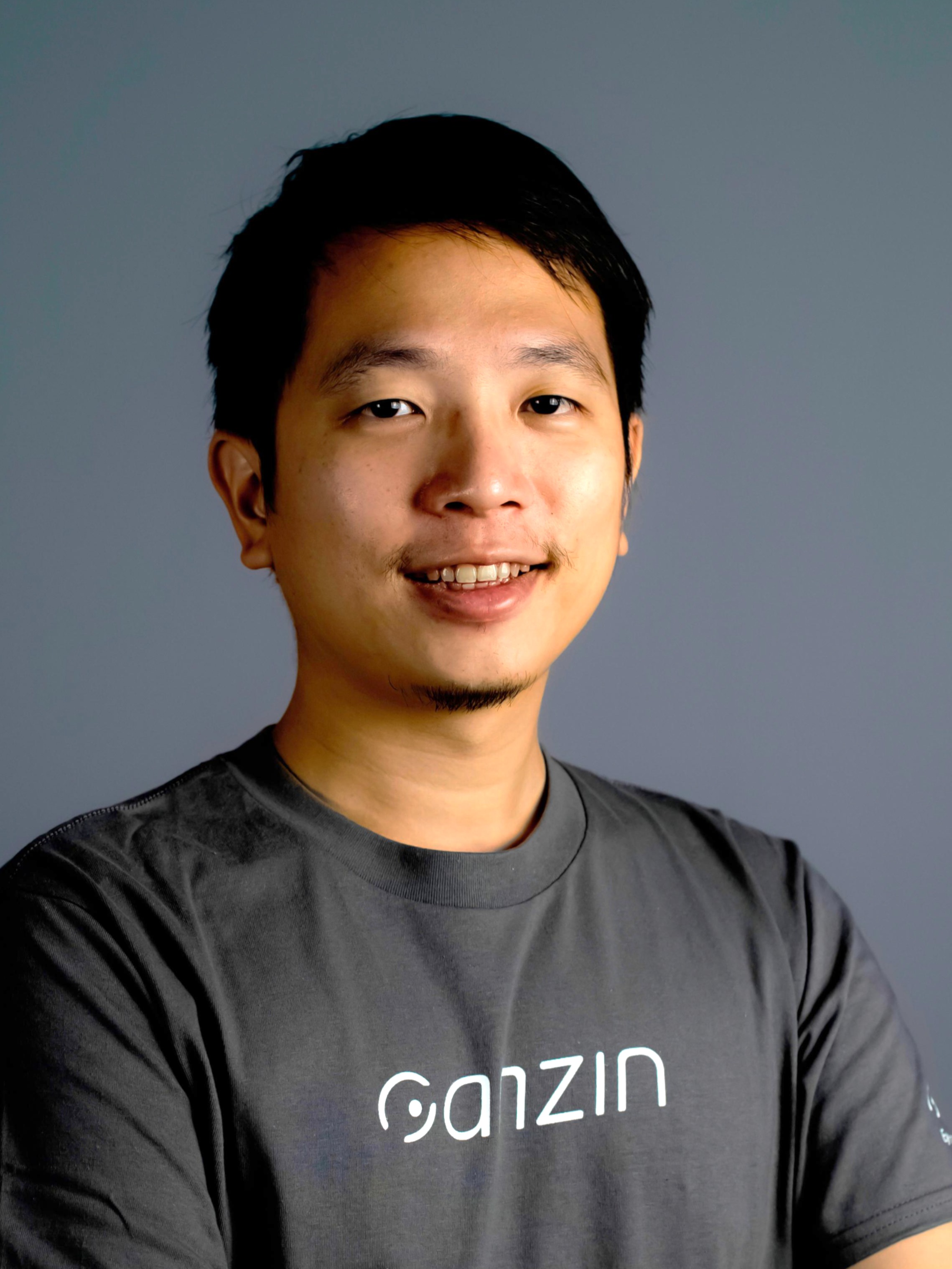}}]{Po-Hsien Yu} received the B.S. degree in mechanical engineering from National Sun Yat-sen University, Kaohsiung, Taiwan, in 2008, and the M.S. degree in electronic engineering from National Yang Ming Chiao Tung University, Hsinchu, Taiwan, in 2016. He is currently a Senior ASIC Design Engineer with Raydium Semiconductor Corporation, Hsinchu, Taiwan, and is pursuing the Ph.D. degree with the Graduate Institute of Electronics Engineering, National Taiwan University, Taipei, Taiwan. His research interests include computer vision, deep learning, federated learning, intelligent surveillance systems, efficient AI algorithms and accelerators, and low-power system-on-chip design.
\end{IEEEbiography}

\begin{IEEEbiography}[{\includegraphics[width=1in,height=1.25in, clip,keepaspectratio]{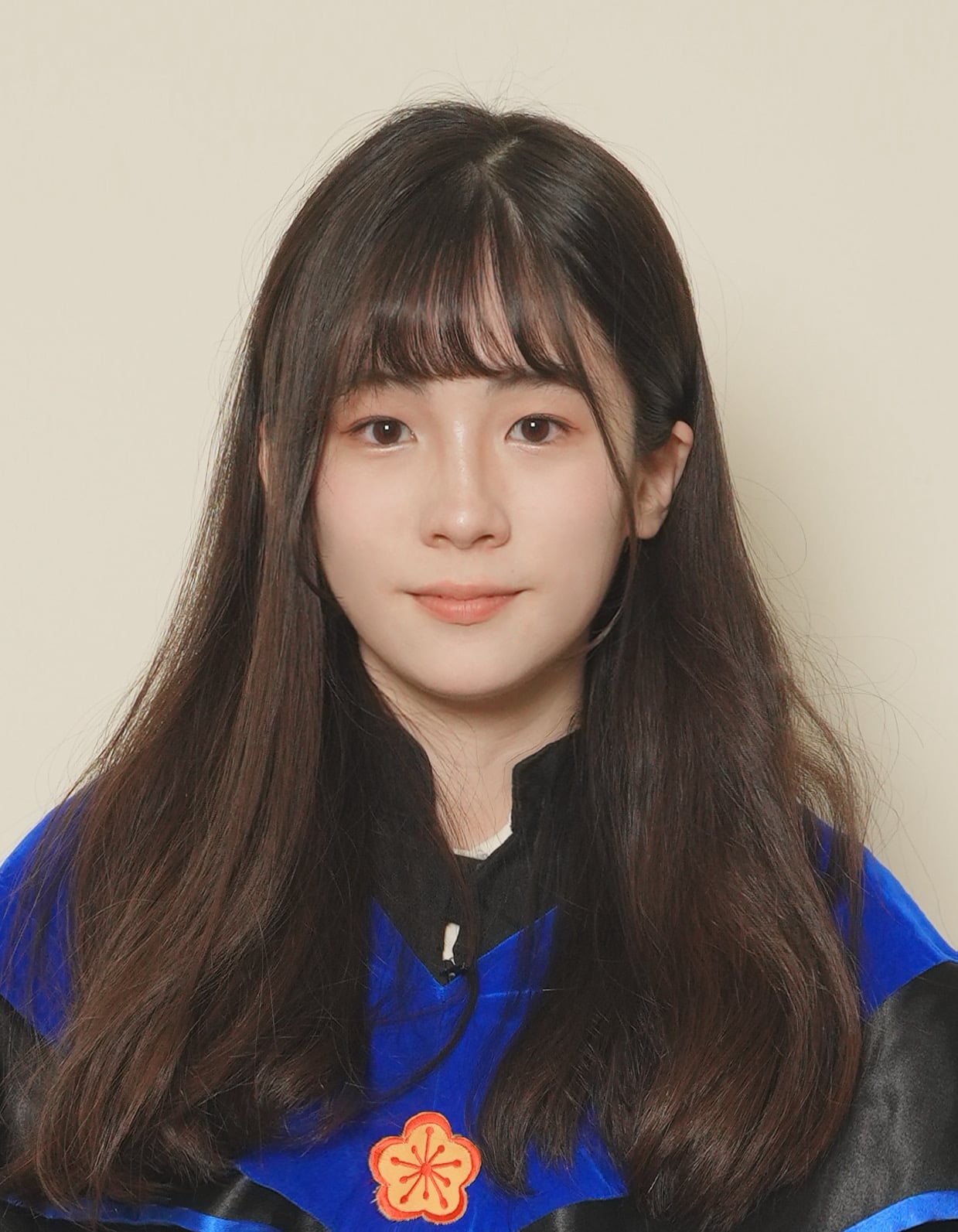}}]{Yu-Syuan Tseng} 
received her B.S. degree in Electrical Engineering from National Central University in 2022. She then obtained her M.S. degree from the Graduate Institute of Electronics Engineering at National Taiwan University in 2025. Her research interests include computer vision algorithms and intelligent surveillance systems.
\end{IEEEbiography}

\begin{IEEEbiography}[{\includegraphics[width=1in,height=1.25in, clip,keepaspectratio]{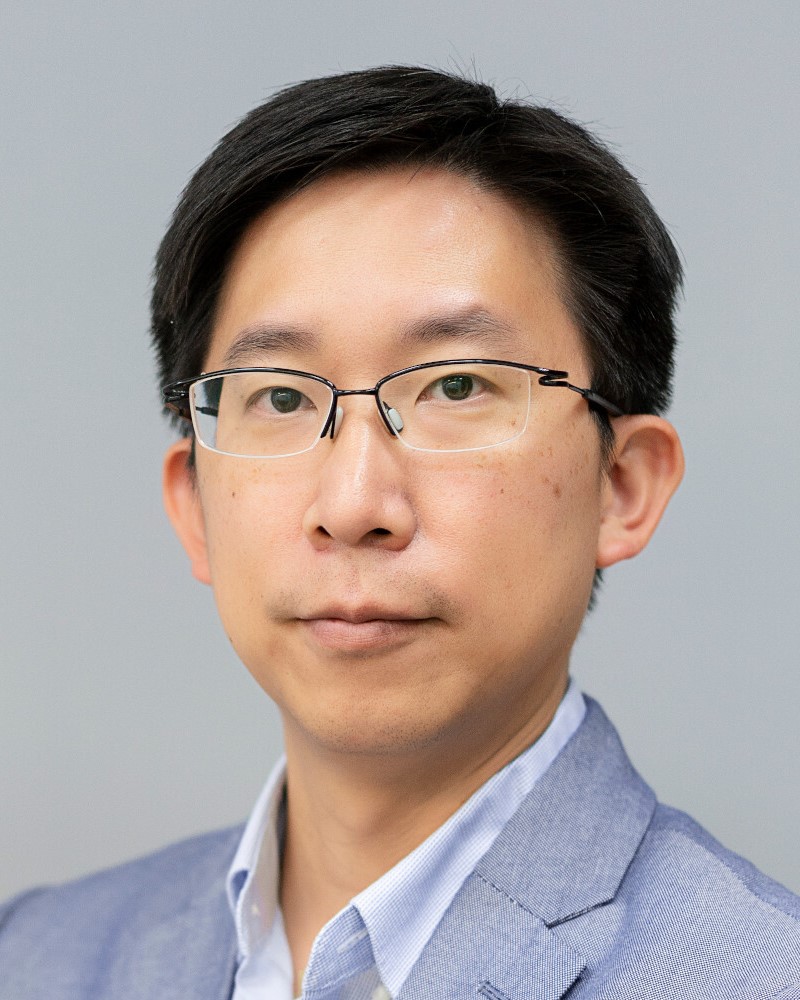}}]{Shao-Yi Chien} (Member, IEEE) received the B.S. and Ph.D. degrees from the Department of Electrical Engineering, National Taiwan University, Taipei, Taiwan, in 1999 and 2003, respectively. During 2003 to 2004, he was a research staff in Quanta Research Institute, Tao Yuan County, Taiwan. In 2004, he joined the Graduate Institute of Electronics Engineering and Department of Electrical Engineering, National Taiwan University, as an Assistant Professor. Since 2012, he has been a Professor. Prof. Chien served as the vice chairperson of the Department of Electrical Engineering, National Taiwan University, in 2013--2016. Since 2018, he has been the founder and CEO of Ganzin Technology, Inc., an eye-tracking solution provider. His research interests include AR/VR, eye tracking, computer vision, real-time image/video processing, video coding, computer graphics, and the associated VLSI and processor architectures.

Dr. Chien served as an Associate Editor for IEEE Transactions on Circuits and Systems for Video Technology in 2009--2016 and Springer Circuits, Systems and Signal Processing (CSSP) in 2009--2015. He also served as an Associate Editor for IEEE Transactions on Circuits and Systems I: Regular Papers in 2012–-2013 and served as a Guest Editor for Springer Journal of Signal Processing Systems in 2008. He also serves on the technical program committees of several conferences, such as ISCAS, ICME, AICAS, SiPS, A-SSCC, and VLSI-DAT.
\end{IEEEbiography}


\end{document}